\documentclass[letterpaper, 10 pt, journal, twoside]{ieeetran}
\usepackage{graphics} % for pdf, bitmapped graphics files
\usepackage{epsfig} % for postscript graphics files
\usepackage{mathptmx} % assumes new font selection scheme installed
\usepackage{times} % assumes new font selection scheme installed
\usepackage{amsmath} % assumes amsmath package installed
\usepackage{amssymb}  % assumes amsmath package installed
\usepackage{listings}
\usepackage{booktabs}
\usepackage{multicol}
\usepackage{minted}
\usepackage{multirow}
\usepackage{stfloats}
\usepackage{threeparttable} 
\usepackage[normalem]{ulem}
\usepackage{tikz}
\usetikzlibrary{trees}
\usepackage{bbding}
\usepackage[hyphens]{url}
\usepackage[colorlinks=True,
            linkcolor=blue,
            citecolor=blue,
            urlcolor=magenta,
            breaklinks=true]{hyperref}
\usepackage{xcolor} % text color
\definecolor{myyellow}{RGB}{255, 200, 0} % 自定义一个比较深的黄色
\definecolor{deepyellow}{RGB}{184, 134, 11} % 深黄色（琥珀色系）
\definecolor{deepgreen}{RGB}{0, 100, 0}    % 深绿色（森林绿系）
\begin{document}
\title{EquiMus: Energy-Equivalent Dynamic Modeling and Simulation of Musculoskeletal Robots Driven by Linear Elastic Actuators}
\author{
Yinglei Zhu$^{1}$, Xuguang Dong$^{1}$, Qiyao Wang$^{1}$, Qi Shao$^{1}$, Fugui Xie$^{1}$, Xinjun Liu$^{1}$, Huichan Zhao$^{1}$%
\thanks{This work was supported by the National Natural Science Foundation of China under Grant No. 52222502 and the Beijing Municipal Natural Science Foundation under Grant No. E2024202287. (\textit{Corresponding author: Huichan Zhao (zhaohuichan@mail.tsinghua.edu.cn)})}
\thanks{All authors are with the Department of Mechanical Engineering, State Key Laboratory of Tribology in Advanced Equipment, Beijing Key Laboratory of Transformative High-end Manufacturing Equipment and Technology, Tsinghua University, Beijing 100084, China.}
}

% Note: use short here
\markboth{Preprint Version. Author Accepted Manuscript, 2025}
{Zhu \MakeLowercase{\textit{et al.}}: EquiMus: Energy-Equivalent Modeling and Simulation of Musculoskeletal Robots} 

\maketitle

\begin{abstract}
Dynamic modeling and control are critical for unleashing soft robots' potential, yet remain challenging due to their complex constitutive behaviors and real-world operating conditions. Bio-inspired musculoskeletal robots, which integrate rigid skeletons with soft actuators, combine high load-bearing capacity with inherent flexibility. Although actuation dynamics have been studied through experimental methods and surrogate models, accurate and effective modeling and simulation remain a significant challenge, especially for large-scale hybrid rigid--soft robots with continuously distributed mass, kinematic loops, and diverse motion modes.

To address these challenges, we propose \textit{EquiMus}, an energy-equivalent dynamic modeling framework and MuJoCo-based simulation for musculoskeletal rigid--soft hybrid robots with linear elastic actuators. The equivalence and effectiveness of the proposed approach are validated and examined through both simulations and real-world experiments on a bionic robotic leg. EquiMus further demonstrates its utility for downstream tasks, including controller design and learning-based control strategies.
\end{abstract}


\section{Introduction}
\IEEEPARstart{B}{io-inspired} robots have been extensively studied and developed in recent years \cite{kim2013soft, iida2016biologically, roderick2021bird, chen2017biologically, hughes2018anthropomorphic}. Among them, articulated musculoskeletal robots feature rigid skeletons and soft artificial muscles, enhancing the robot's load capacity and intrinsic adaptability \cite{feng2024experimental, deimel2016novel, tolley2014resilient}. In these systems, elastic actuators (EAs) boast high energy density and inherent compliance \cite{Dong2025}. Dynamic modeling and simulation of these systems lay important foundations for their design, control \cite{zhang2019modeling}, and data generation \cite{della2023model}.

However, classical rigid–body models do not capture configuration-dependent mass redistribution, large elastic strains with damping, or loop-closure constraints in rigid–soft hybrids \cite{graule2021somo}. In practice, soft actuators often have non-negligible mass in load-bearing limbs, so their inertia must be modeled alongside actuation. As the configuration changes, composite inertia varies, and kinematic loops frequently arise in such systems, both of which complicate the dynamics and make per-robot Lagrangian derivations labor-intensive and hard to generalize.

Meanwhile, rigid-body robotics has advanced rapidly, fueled by physics engines and graphics computation \cite{kober2013reinforcement}. Simulators such as PyBullet \cite{coumans2021pybullet}, MuJoCo (Multi-Joint dynamics with Contact) \cite{todorov2012mujoco} and Isaac Gym \cite{makoviychuk2021isaac} offer standardized environments well suited for data-driven control and design \cite{miki2022learning, rudin2022learning, li2021reinforcement}. Although MuJoCo can mimic viscoelastic behavior \cite{wang2022myosim, caggiano2022myosuite}, it omits actuator inertia, limiting accurate simulation of coupled rigid--soft dynamics.
\begin{figure}[htbp]
    \centering
    \includegraphics[width=1\linewidth]{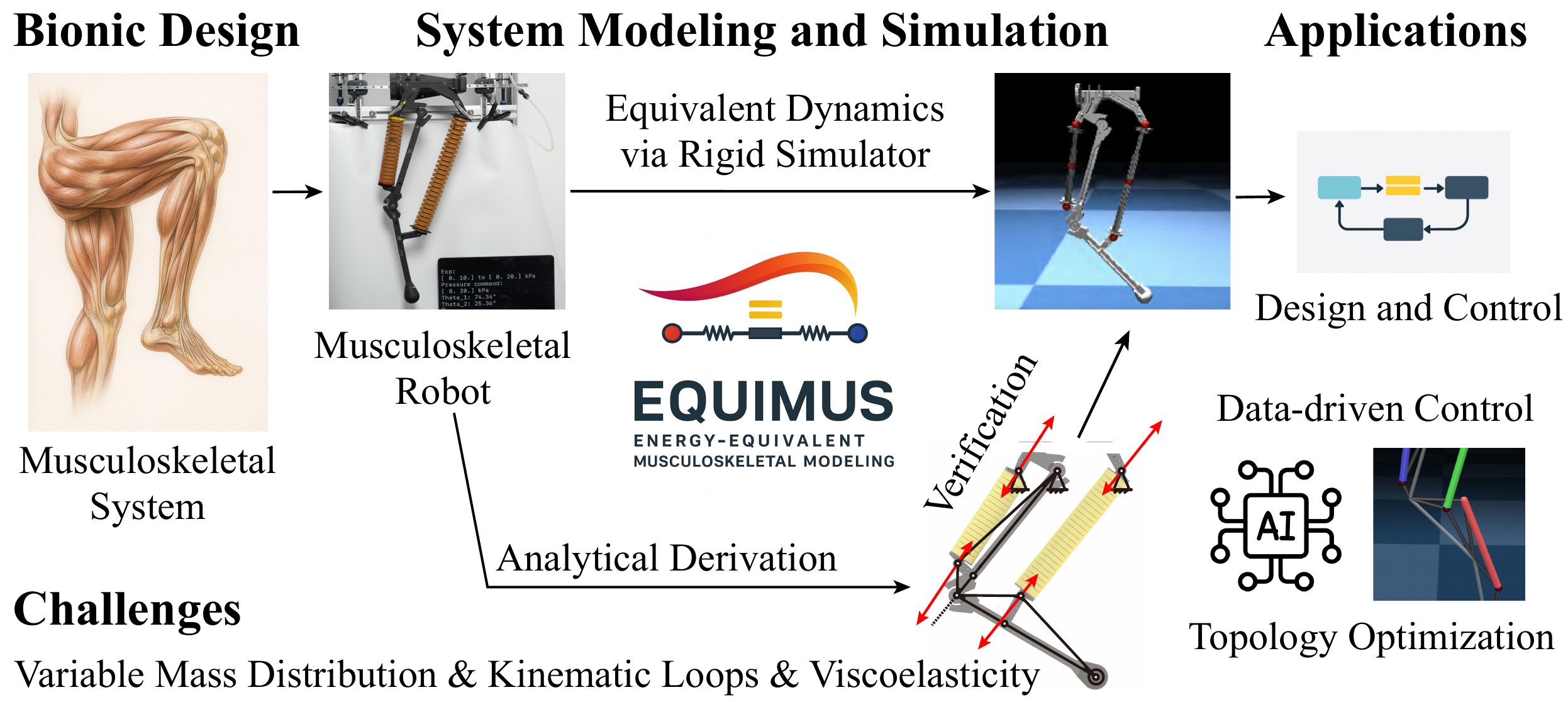}
    \caption{Overview and motivation of the proposed energy-equivalent modeling framework EquiMus.}
    \label{pic-overview}
\end{figure}
To bridge these gaps, we present \textit{EquiMus}\footnote{Code and derivation are available at 
\href{https://github.com/fly-pigTH/EquiMus}
{\textnormal{https://github.com/fly-pigTH/EquiMus}}}, an energy-equivalent modeling and MuJoCo-based simulation framework for rigid–soft hybrid robots with linear elastic actuators. As shown in Fig.~\ref{pic-overview}, EquiMus tackles three challenges: (i) the actuator inertia and variable mass distribution during motion; (ii) loop-closure within musculoskeletal chains; (iii) plug-and-play integration with model-based and learning-based controllers. The main contributions of this work are summarized as:
\begin{enumerate}
    \item \textbf{Modeling}: a compact energy-equivalent lumped-mass formulation that maps elastic actuator dynamics to discrete rigid-body elements, preserving energies and matching the virtual work of damping and actuation;
    \item \textbf{Implementation}: a MuJoCo realization with loop-closure constraints, smooth elastic actuator dynamics, and out-of-the-box reinforcement learning (RL) compatibility;
    \item \textbf{Validation and Applications}: experiments on a pneumatic leg with fluidic elastomer actuators (FEAs), show close sim-to-real agreement and enable PID auto-tuning, model-based control, and RL-based control.
\end{enumerate}
\begin{table*}[!b]
\centering
\caption{Comparison of Different Simulation Frameworks}
\label{tab-simulators}
\renewcommand{\arraystretch}{1.2}
\begin{threeparttable}
\begin{tabular}{l@{\hspace{6pt}}c@{\hspace{6pt}}c@{\hspace{6pt}}c@{\hspace{6pt}}c@{\hspace{6pt}}c@{\hspace{6pt}}c}
\toprule
\textbf{Framework} & \textbf{Modeling Approach} & \textbf{Physics Complexity} & \textbf{Kinematic Loop} & \textbf{R--S Hybrid Robots} & \textbf{DMR} \\
\midrule
PyBullet/Gazebo     & rigid-body        & low    & $\times$ & $\times$ & $\times$ \\
Webots       & rigid-body        & low     & $\times$ & $\times$ & $\times$\\
Elastica     & Cosserat rods     & medium   & $\times$ & $\times$ & $\checkmark$\\
SOFA         & FEM, finite element method             & high    & $\checkmark$ & $\checkmark$ & $\checkmark$ \\
SoMo/SoMoGym         & rigid-body        & low     & $\times$ & $\mathbf{\checkmark}$ & $\checkmark$\\
SoftManiSim  & rigid-body + Cosserat & medium & $\times$ & $\checkmark$ & $\checkmark$\\
OpenSim       & rigid-body + muscle model  & medium-high & $\checkmark$ & $\checkmark$ & $\times$\\
MuJoCo/MyoSim       & rigid-body + muscle model & medium    & $\checkmark$ & $\checkmark$ & $\times$\\
\textbf{EquiMus*}    & \textbf{rigid-body equivalence} & \textbf{low} & \CheckmarkBold & \CheckmarkBold & \CheckmarkBold\\
\bottomrule
\end{tabular}
\begin{tablenotes}[flushleft]
\setlength{\labelsep}{0pt}
\setlength{\leftmargin}{0pt}
\footnotesize
\item R--S = rigid--soft; DMR = dynamic mass redistribution (in this work).
\end{tablenotes}
\end{threeparttable}
\end{table*}

\section{Literature Review}

\subsection{Dynamic Modeling of Soft Robots}

Modeling soft robots remains challenging due to inherent continuum properties, nonlinear material behaviors, and high-dimensional configuration spaces. Among them, articulated soft robots, which are rigid--soft hybrids with discrete links and soft actuation, represent a distinct subclass that requires tailored modeling considerations. Existing methods can be categorized into four mainstream approaches \cite{Armanini23}:

\textbf{Continuum mechanics models} (e.g., Cosserat rod and nonlinear Euler-Bernoulli beam formulations \cite{till2019real}) deliver high fidelity for slender, continuously deformable structures but incur prohibitive computation, making them unsuitable for real-time control in articulated hybrids.

\textbf{Geometrical models} (e.g., piecewise-constant-curvature \cite{webster2010design}) approximate deformations via low-dimensional curves, trading expressiveness for computational speed for continuum manipulators, but limited for multi-link hybrids.

\textbf{Discrete material models} condense mass and compliance into discrete elements \cite{habibi2020lumped}, balancing accuracy and efficiency---suited for articulated hybrids with distributed elasticity.

\textbf{Surrogate models} apply neural networks (NNs) to model nonlinear behaviors \cite{giorelli2015neural} but demand extensive training data, relying on fast simulators for dataset generation. 

Accordingly, we adopt a discrete (lumped-mass) approach, because it preserves actuator inertia at real-time rates and strikes the desired fidelity-efficiency balance.

\subsection{Simulation of Soft Robots}
Hand-crafting dynamic models for soft/rigid–soft systems is costly, and coupling them with modern controllers (e.g., reinforcement learning) is labor-intensive. With advances in biomechanics and graphics, simulators have become the default for design, analysis, and data generation.

Early PyBullet-based, continuum-oriented efforts such as SoMo \cite{graule2021somo} and SoMoGym \cite{graule2022somogym} approximate soft manipulators via segment discretization with beam theory. These methods capture quasi-static behavior effectively but lack full dynamics. Moreover, because the Unified Robot Description Format (URDF) assumes open chains, they cannot represent the loop closures \cite{chignoli2024urdf} that are common in articulated musculoskeletal robots. More physics-rich simulators (e.g., SoftManiSim \cite{kasaei2024softmanisim} and Jitosho's framework \cite{jitosho2023reinforcement}) incorporate advanced Cosserat/beam theory and demonstrate RL-based locomotion. However, they are typically confined to centimeter-scale tasks and incur significant computational cost, limiting real-time control studies. FEM-based SOFA (Simulation Open Framework Architecture) \cite{faure2012sofa} offers high physical fidelity for soft tissues and continuum structures but is likewise heavy for real-time applications.

By contrast, MuJoCo \cite{todorov2012mujoco} has become a preferred platform for articulated systems and has been largely used for model-based/data-driven control and sim-to-sim testing of rigid-body robots \cite{li2024reinforcement, andrychowicz2020learning}. Its XML format supports kinematic loops and musculoskeletal simulations \cite{wang2022myosim, caggiano2022myosuite}. However, MuJoCo's actuators are still modeled as massless elements, limiting its ability to capture dynamic mass redistribution (DMR)-i.e., changes in the mass distribution of embedded soft actuators during motions. OpenSim \cite{delp2007opensim, seth2011opensim, seth2018opensim} shares this limitation.

Classic simulators are summarized in Table \ref{tab-simulators}. It highlights key capabilities such as support for kinematic loops and the ability to handle dynamic mass redistribution. In summary, off-the-shelf simulators still have limitations in realizing the dynamic modeling and simulations for musculoskeletal robots.

\section{Methods}

\subsection{Basic Assumptions}
An elastic actuator, illustrated in Fig.~\ref{pic-lumped}, has endpoints A, B, mass $m$, stiffness $k$, damping $c$, rest length $l_0$, current length $l$, and driving force $\mathbf{F}$. The actuator is assumed to act as a driving unit in the rigid--soft hybrid robot and follow basic conditions:
\begin{itemize}
    \item \textbf{Unidirectional and Uniform Deformation}: The actuator, with a negligible radial size, is modeled as a one-dimensional linear element with a uniform mass distribution, with a linear density, denoted as $\rho=m/l$.
    \item \textbf{Axial Driving Forces}: The actuator is subjected to a pair of equal and axial forces, denoted as $\mathbf{F}$, applied at endpoints $\mathrm{A}$ and $\mathrm{B}$ along its axis.
    \item \textbf{Spring-Mass-Damper System}: With one end fixed, we assume that the actuator behaves as a second-order system with constant elasticity and viscosity \cite{reynolds2003modeling} \cite{mizuno2011spring}.
\end{itemize}

\subsection{Energy-Based Perspective on Dynamic Modeling}

The robot dynamics are formulated using the vector form of the Lagrangian equation,
\begin{align}
    (\frac{d}{dt} \frac{\partial }{\partial \dot{\mathbf{q}}} - \frac{\partial }{\partial \mathbf{q}}) (L_{\text{EA}}+L_{\text{other}}) = \mathbf{Q}_{\text{EA}} + \mathbf{Q}_{\text{other}}
\end{align}
where $L$, $\mathbf{q}$, and $\mathbf{Q}$ denote the Lagrangian, generalized coordinates, and generalized forces respectively. We decompose $L$ and $\mathbf{Q}$ into contributions from elastic actuators (EA) and rigid structures (other). $L_{\text{other}}$ and $\mathbf{Q}_{\text{other}}$ depend on $\mathbf{q}$, its derivative $\dot{\mathbf{q}}$, and external inputs. From an energy perspective, if the energy and forces of the elastic actuator can be discretized with rigid-body equivalents, the overall dynamics remain invariant, regardless of the specific configuration and type of soft actuators. 

\subsection{Energy-Equivalent Model}

Inspired by the lumped mass method \cite{Armanini23, habibi2020lumped}, we discretize the actuator into an assumed energy-equivalent mass-spring-damper array. This formulation integrates the dynamics of the soft actuator into a multi-rigid-body representation, accounting for the elastoplastic behavior, variable mass distribution, and kinematic loops. The theoretical proof is provided through constructive modeling and parameter derivation.

Let $\vec{r}_\mathrm{A}$ and $\vec{r}_\mathrm{B}$ denote the position vectors of endpoints A and B in the global (inertial) frame, respectively. Suppose $N$ mass points are distributed along the elastic actuator at positions $\vec{x}_1, \ldots, \vec{x}_N$, with corresponding masses $m_1, \ldots, m_N$. Each adjacent pair of points is connected by a linear actuator characterized by stiffness $k_i$, damping $c_i$, original length $l_{i0}$, and internal driving force $F_i$, where $i = 1,\ldots,N{-}1$. For convenience, $\vec{x}_i$ is reduced to 1D coordinate $\xi_i$ along the actuator’s length, such that. $\vec{x}_i(\xi_i)= (1-\xi_i) \vec{r}_\mathrm{A} + \xi_i \vec{r}_\mathrm{B}$. In this section, energy quantities in the equivalent model are denoted by $\hat{\cdot}$.

\subsubsection{Equivalence of Gravitational Potential Energy}
The gravitational potential energy of the actuator and its equivalent rigid-body model is given by
\begin{align}
    V_\mathrm{g} &= \frac{1}{2}m\mathrm{g}\vec{k}\cdot(\vec{r}_\mathrm{A} + \vec{r}_\mathrm{B}) \label{eq-Vg}\\
    \hat{V}_\mathrm{g} 
    &= \sum\nolimits_{i=1}^{N} m_i \mathrm{g} \vec{k}\cdot \vec{x}_i = \sum\nolimits_{i=1}^{N} m_i \mathrm{g}\vec{k}\cdot \left[(1-\xi_i) \vec{r}_\mathrm{A} + \xi_i \vec{r}_\mathrm{B}\right]\label{eq-EQVg}
\end{align}
where $\vec{k}$ denotes the unit vector of the ground frame, oriented opposite to gravity. 

\subsubsection{Equivalence of Kinetic Energy}
Assuming uniform linear elongation, the velocity of the microelement $\mathrm{d}x$ at position $x$ along the actuator varies linearly, $\vec{v}(x) = \vec{v}_\mathrm{A} \cdot \frac{l - x}{l} + \vec{v}_\mathrm{B} \cdot \frac{x}{l}$. By integrating over the entire length, the total kinetic energy of the actuator is given by:
\begin{align}
   T &= \int_0^{l} \frac{1}{2} v^2(x) \frac{m}{l} \ \mathrm{d}x = \frac{1}{6}m(\vec{v}_\mathrm{A}\cdot \vec{v}_\mathrm{B}+v_{\mathrm{A}}^2+v_{\mathrm{B}}^2) \label{eq-T}\\
   \hat{T} &=\sum_{i=1}^{N} \frac{1}{2} m_i \|(1-\xi_i) \vec{v}_\mathrm{A} + \xi_i \vec{v}_\mathrm{B}\|^2.\label{eq-EQT}
\end{align}
\begin{figure}[h]
    \centering \includegraphics[width=0.9\linewidth]{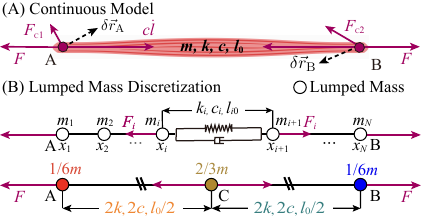}
    \caption{Schematic diagram of the soft actuator and its lumped mass distribution. 
    The actuator is subjected to gravity, driving forces $F$, elastic forces, viscous resistance $c\dot{l}$ at endpoints, and constraint forces $F_\mathrm{c}$ from the rigid skeletons. $\delta\vec{r}$ denotes the virtual displacement.
    Since $F_\mathrm{c}$ is internal to the system, it is excluded from the Lagrangian dynamics. Driving forces and viscous forces are defined as positive as shown in figure. }
    \label{pic-lumped}
\end{figure}
Let $\mu_i =  \frac{m_i}{m}$. Energy-equivalence requires $\hat{V}_\mathrm{g}=V_\mathrm{g},\ \hat{T}=T$. By substituting into Eqs.~\eqref{eq-Vg}, \eqref{eq-EQVg}, ~\eqref{eq-T} and \eqref{eq-EQT}, 
%{\setlength{\arraycolsep}{3pt}  % 减小列间距
\begin{align}
    \left.
    \begin{aligned}
        \hat{T} &= T \\
        \hat{V}_\mathrm{g} &= V_\mathrm{g}
    \end{aligned}
    \right\}
    \Longleftrightarrow
    \begin{bmatrix}
    1 &  \cdots & 1 \\
    \xi_1  & \cdots & \xi_N \\
    \xi_1^2  & \cdots & \xi_N^2
    \end{bmatrix}
    \cdot
    \begin{bmatrix}
    \mu_1 \\
    \vdots \\
    \mu_N
    \end{bmatrix}
    =
    \begin{bmatrix}
    1 \\
    1/2 \\
    1/3
    \end{bmatrix}.
\end{align}%}
The linear system admits solutions only if $N\geq3$. When $N=3$, the coefficient matrix reduces to a convertible 3rd-order Vandermonde matrix for any ($\xi_1, \xi_2 , \xi_3$) that meets $\xi_1 \neq \xi_2 \neq \xi_3$, ensuring the existence of a solution. That is the necessary condition of a rigid discrete equivalent model. 

Without loss of generality, we set $\xi_1 = 0$ and $\xi_3 = 1$, to anchor the ends of the actuator to the skeleton. To achieve better symmetry and a simpler constraint of length, we select $\xi_2 = \frac{1}{2}$, resulting in the solution $\textcolor{red}{\mu_1 = \frac{1}{6}}$, $\textcolor{deepyellow}{\mu_2 = \frac{2}{3}}$, and $\textcolor{blue}{\mu_3 = \frac{1}{6}}$. This solution corresponds to Eq.~\eqref{eq-V_split} and \eqref{eq-Tsquare}. Additionally, the midpoint is the optimal choice for interpolation accuracy according to Simpson's interpolation theory.
\begin{align}
    V_\mathrm{g} &= m\mathrm{g} \vec{k}\cdot \left[\textcolor{red}{\frac{1}{6}}\vec{r}_\mathrm{A} + \textcolor{deepyellow}{\frac{2}{3}}(\frac{\vec{r}_\mathrm{A} + \vec{r}_\mathrm{B}}{2}) + \textcolor{blue}{\frac{1}{6}}\vec{r}_\mathrm{B}\right] =\hat{V}_\mathrm{g} \label{eq-V_split}\\
     T &=\frac{1}{2}m\left[\textcolor{red}{\frac{1}{6}} v_\mathrm{A}^2 +\textcolor{deepyellow}{\frac{2}{3}} (\frac{\vec{v}_\mathrm{A}+\vec{v}_\mathrm{B}}{2})^2 + \textcolor{blue}{\frac{1}{6}} v_\mathrm{B}^2\right] = \hat{T} \label{eq-Tsquare}
\end{align}

\subsubsection{Generalized Forces and Elastic Potential Energy}
Under the above assumptions, and by analogy with series rules for springs and dampers, dynamic equivalence holds if the following conditions are met:
\begin{equation}
\setlength{\arraycolsep}{3pt}
\renewcommand{\arraystretch}{1.1}
\left\{
\begin{aligned}
    \textcolor{orange}{F_1} &= \textcolor{teal}{F_2} = F, &
    \textcolor{orange}{c_1} &= \textcolor{teal}{c_2} = 2c, \\
    \textcolor{orange}{k_1} &= \textcolor{teal}{k_2} = 2k, &
    \textcolor{orange}{l_{10}} &= \textcolor{teal}{l_{20}} = l_0/2.
\end{aligned}
\right.
\end{equation}
The elastic potential energy and virtual work brought by generalized forces (driving force $F$ and the damping force $c \dot{l}$) matches:
\begin{align}
   \hat{V}_\mathrm{e} &= \frac{1}{2} \textcolor{orange}{k_1}(l/2 - \textcolor{orange}{l_{10}})^2 + \frac{1}{2} \textcolor{teal}{k_2}(l/2 - \textcolor{teal}{l_{20}})^2 \notag \\
   &=2 \times \frac{1}{2} \times 2k \left[({l-l_0})/2\right]^2 
   = \frac{1}{2} k \left[(l-l_0)\right]^2 = V_\mathrm{e} \label{eq-Ve}\\
    \delta \hat{W} &=(\textcolor{orange}{F_1}-\textcolor{orange}{c_1}\dot{l}/2)\vec{e}_{\mathrm{CA}}\cdot\delta \vec{r}_{\mathrm{CA}} + (\textcolor{teal}{F_2}-\textcolor{teal}{c_2}\dot{l}/2)\vec{e}_{\mathrm{CB}}\cdot\delta \vec{r}_{\mathrm{CB}} \notag \\
    &=2 \times (F-c\dot{l})\vec{e}_{\mathrm{BA}}\cdot 1/2\delta \vec{r}_{\mathrm{BA}} = \delta{W} \label{eq-W}
\end{align}
where $\vec{e}_{\mathrm{AB}}$ denotes the unit vector from A to B. According to the principle of virtual work, the equivalence in virtual work implies that of generalized forces. Consequently, the transformation constructed above preserves all physical terms, resulting in full equivalence of the system dynamics.
\begin{figure*}[t]
\centering
\begin{tikzpicture}[
  level distance=10mm,        % 每一层之间的垂直距离
  sibling distance=42mm,      % 同层节点的水平间距
  every node/.style={
    draw=black, thick,
    rectangle, rounded corners,
    align=left,
    font=\scriptsize,
    fill=gray!10,
    text width=3.2cm           % 控制节点宽度，方便多行显示
  },
  edge from parent/.style={draw=black, thick, -latex}
]

\node (b1) { \textbf{Body: UpperSkeleton} }
    % ---- Link1 ----
    child {node { \textbf{Body: muscleX\_uLink}}
        child{node { \textbf{Joint: muscleX\_uRotJoint}}}
        child{node { \textbf{Geom: muscleX\_uGeom}\\mass = $m/6$}}
        child{node { \textbf{Body: muscleX\_mLink}}
            child {node(j1) {\textbf{Joint: muscleX\_mSlideJoint}\\
                type = slide 	
                stiffness = $2k$ \\
                damping = $2c$ 
                springref = $l_0/2$ }
            }
            child {node { \textbf{Geom: muscleX\_mGeom} \\ mass = $2/3m$ }}
            child {node (b3) { \textbf{Body: muscleX\_lLink}}
                child {node(j2) {\textbf{Joint: muscleX\_lSlideJoint}\\
                    type = slide 
                    stiffness = $2k$ \\
                    damping = $2c$
                    springref = $l_0/2$ }}
                child {node { \textbf{Geom: muscleX\_lGeom} \\ mass = $m/6$ }}
                child [missing] {}
            }
        }
    }
    child { node[font=\Large, align=center] {...}  % 省略号节点
        child [missing] {}
        child [missing] {}
        child {node (b2) { \textbf{Body: LowerSkeleton}}}
    };
% 连接线，无边框
\draw[->, thick, black, dashed]
    (j1.south) to[bend left=-50] node[draw=none, fill=none, pos=0.6, xshift=25pt, yshift=5pt, above, text=black]{\textbf{Equality: joint}} (j2.west);
\draw[->, thick, black, dashed]
    (b2.east) to[bend left=50] node[draw=none, fill=none, pos=0.5, xshift=-25pt, yshift=7pt, below, text=black]{\textbf{Equality: connect}} (b3.east);

\end{tikzpicture}
\caption{MJCF hierarchical structure of the EquiMus model, showing body–joint–geom relationships and key attributes. The “...” node denotes the remaining rigid skeleton structure, omitted here for clarity. Dashed arrows indicate \texttt{<equality>} constraints, including joint equality and body connection.}
\label{fig:mjcf-tree}
\end{figure*}
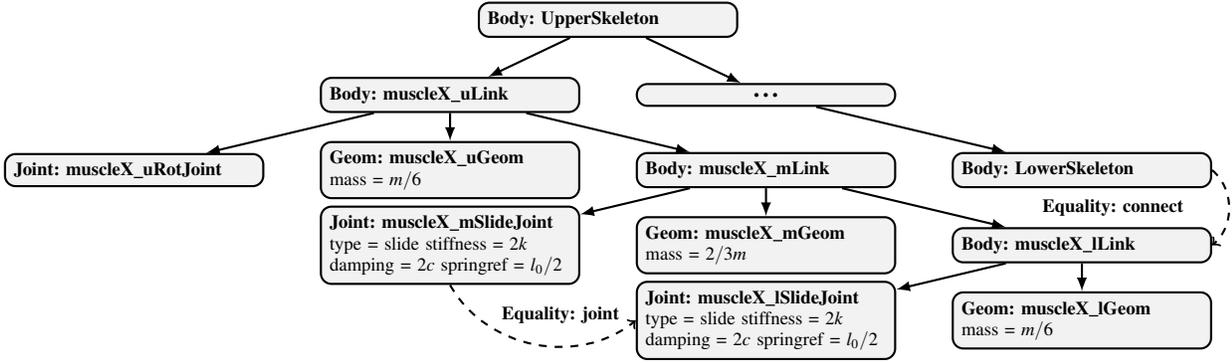

\subsection{Summary}

Shown in Fig.~\ref{pic-lumped}, the dynamic model of the linear elastic actuator can be equivalently represented by a discrete mass system. The method follows a 3-2-1 approach\label{sec:321approach}:
\begin{itemize}
    \item \textbf{3 Mass Points}: The actuator is discretized into three mass points---two fixed at each end and one at the midpoint---with $\frac{1}{6}m$, $\frac{1}{6}m$, and $\frac{2}{3}m$ respectively.
    \item \textbf{2 Linear Actuators}: Two motors connect the masses, each with stiffness $2k$, damping coefficient $2c$, rest length of $l_0/2$, and driving forces $F$.
    \item \textbf{1 Constraint}: An equality constraint is implicitly enforced through the above conditions, ensuring that the elongation of both actuator segments remain identical.
\end{itemize}
Our method achieves a balanced trade-off between physical fidelity and computational efficiency.
On one hand, the equivalent model is rigorously constructed based on energy equivalence, carefully accounting for the variable mass distribution, thereby preserving high accuracy. On the other hand, although the soft actuator is discretized into a multi-rigid-body representation, the number of equivalent elements and constraints remains comparable to that of the original actuator, substantially reducing computational overhead.

\subsection{Software Implementation}
We choose MuJoCo \cite{todorov2012mujoco} as simulation platform due to its effective support for kinematic loops. The energy-equivalent implementation of the linear elastic actuator is illustrated in Fig.~\ref{fig:mjcf-tree}. Parameters are configured with a 1:4:1 \verb|<mass>| distribution, double \verb|<stiffness>| and \verb|<damping>| coefficients, and half \verb|<springref>| (rest length). An \verb|<equality>| on joint constraint maintains equal actuator lengths, keeping the middle mass point fixed at the geometric midpoint.

\section{Experiments and Results}
\subsection{Experimental Platform of rigid--soft Hybrid Robotic Leg}
\subsubsection{Overview of System}
We validate EquiMus on a pneumatic rigid--soft robotic leg \cite{Dong2025}. The leg, shown in Fig.~\ref{fig-mujocomodel}, consists of a base, hip joint, thigh, knee joint, calf, a mono-joint actuator (MAA), and a bi-joint actuator (BAA). Rigid links are 3D printed from nylon, and linear elastic actuators are cast from polyurethane rubber. The overall system includes:
\begin{itemize}
    \item \textbf{Actuation:} Two pneumatic artificial muscles (PAMs) are driven by proportional pressure valves (VPPE-3-1/8-6-010, Festo). Each actuator is commanded in the range of 0--50 kPa \cite{Dong2025} (air supply up to 600 kPa).
    \item \textbf{Sensoring:} Joint motion is tracked by a motion-capture system (Optitrack, Natural Point) at 120 Hz.
    \item \textbf{Control:} A ROS2-based PC receives state estimates and outputs valve pressure commands via DACs.
\end{itemize}
The actuators' physical parameters are summarized in Table~\ref{tab:actuator}. 
\begin{table}[h]
    \centering
    \caption{Actuator Properties} \label{tab:actuator}
    \begin{tabular}{cccccc}
        \toprule
        Actuator & Mass & Len. & Damp. & Stiff. & Area \\
        & ($\text{kg}$) & ($m$) & (Ns/m) & (N/m) & ($\mathrm{m}^{2}$) \\
        \midrule
        MAA & $0.1865$ & $0.1744$ & 10.8 & 367.8 & $6.54 \times 10^{-4}$ \\
        BAA & $0.2727$ & $0.2536$ & 11.3 & 291.8 & $6.37 \times 10^{-4}$ \\
        \bottomrule
    \end{tabular}
\end{table}

\subsubsection{Kinematic Model of Robotic Leg}
We consider two phases: swing (base fixed) and stance (foot in contact). Inputs are actuator pressures $\mathbf{P}=[P_1,P_2]$, which generate axial forces $\mathbf{F}=[F_1,F_2]=[S_1P_1,S_2P_2]$ via the effective cross-sectional areas $S_i$. The outputs are joint angles $\mathbf{q}=[\theta_1,\theta_2]$ of the hip and knee.

\begin{figure*}[t]
\centering
\includegraphics[width=0.99\linewidth]{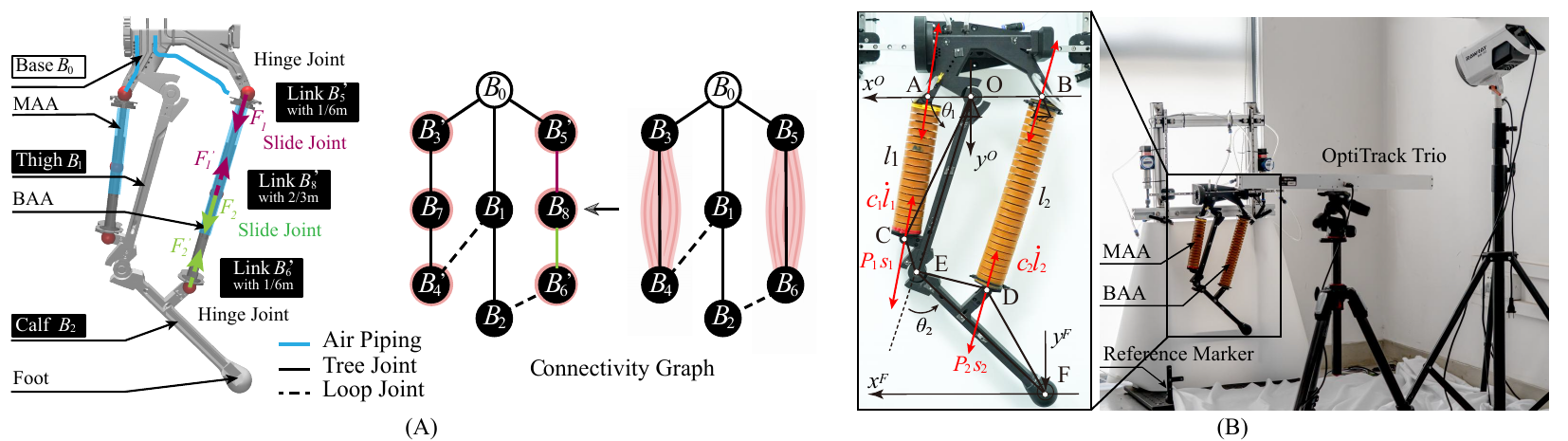}
% \vspace{-5pt}
\caption{
    Comparison between simulation and physical implementation of the robotic leg system. 
    (A) MuJoCo-based energy-equivalent model of the robotic leg, illustrating the complete structure and the corresponding connectivity graph. Black nodes represent rigid links; black solid lines indicate tree joints; black dashed lines denote loop-closure joints (loop joints). Two blue lines represent pneumatic tubing. For clarity, the MAA module is omitted, and the focus is on the BAA structure.
    (B) Physical experimental setup, including the coordinate frame of the bionic robotic leg with pneumatic artificial muscles and the OptiTrack motion capture system. A simplified force analysis diagram is also included to aid understanding of the joint loading conditions.
} \label{fig-mujocomodel}
\end{figure*}

For swing phase, a coordinate system is established as shown in Fig. \ref{fig-mujocomodel} (B), with the origin $\mathrm{O}$ located at the hip joint. The positions and velocities of all key points, $\mathrm{O}, \mathrm{A}, \ldots \mathrm{H}$, as well as the lengths of the two actuators, $l_1, l_2$, can be computed analytically through forward kinematics.

For the stance phase, by choosing the foot position as the origin of the reference frame, the kinematic model can be reformulated via a coordinate transformation $\mathbf{r}^{\mathrm{F}} = \mathbf{r}^{\mathrm{O}} - \overrightarrow{\mathrm{OF}}$.

\subsection{Implementation of Dynamic Simulations}
\subsubsection{Theoretical Dynamic Model and Simulation}

Based on Eqs.~\eqref{eq-Vg}, \eqref{eq-T}, \eqref{eq-Ve}, and \eqref{eq-W}, we derived the dynamics of the robotic leg via the Euler--Lagrange equation, obtaining $\dot{\mathbf{q}}=f_{\text{theo}}(\mathbf{q},t)$. Numerical simulation was implemented in MATLAB/SymPy using the \texttt{ode45} solver, which serves as the theoretical ground truth. 

\subsubsection{Software Pipeline}

The topology transformation and the schematic diagram of the mechanism are shown in Fig.~\ref{fig-mujocomodel}. We implemented the equivalent model of the robotic foot through a systematic process. A URDF file (with loop closures opened) was converted to MJCF, and the 3-2-1 structure (Sec.~\ref{sec:321approach}) was added with the necessary bodies/joints and constraints. Equality constraints were introduced to enforce the actuator lengths and their connections to the skeleton. The simulator runs at 141.9$\times$ real time (mean wall-clock step 0.0071~ms; $\Delta t=0.005$~s), sufficient for real-time control and repeated simulation experiments.

\subsection{Verification in Simulation}

\subsubsection{Static Equivalence}
For each static state $\mathbf{q}$ sampled in the workspace 
$\Omega=\{(\theta_1,~\theta_2)\mid \theta_1\in[\pi/6, 2\pi/3],~\theta_2\in[0,\pi/2]\}$
we compared equilibrium configurations of the theoretical and equivalent models under identical balancing force inputs $\mathbf{F} = f(\mathbf{q})$. Both models shared the same parameters $(l_{10}, l_{20}, k_1, k_2)$ with small perturbations, and $c_1=c_2\approx10~\mathrm{Ns/m}$. Table~\ref{tab:error_summary} summarizes errors over 10,000 trials: RMSEs (root-mean-square error) for $\theta_1$ is $<0.001$~rad and for $\theta_2$ is $<0.06$~rad, indicating close static agreement.

\subsubsection{Dynamic Equivalence}
\paragraph{Swing Phase}
We uniformly sampled $(\mathbf{q},~\mathbf{q}^{\prime})\in\Omega\times\Omega$, computed $\mathbf{F}=f(\mathbf{q})$ and $\mathbf{F}^{\prime}=f(\mathbf{q}^{\prime})$, and finally
recorded the step response from the static state with $\mathbf{F}$ to that of $\mathbf{F}^{\prime}$. Of 10,000 trials, 2,407 valid trajectories remained after discarding joint-limit violations.

\paragraph{Stance Phase}
To test the generalization of our method, we change the working condition to stance phase. The init state is chosen as standing still $(\theta_1 = \pi/2, \theta_2=0)$, the impulse force $\mathbf{F^{\prime}}$ as $(10\ \mathrm{N}, 10\ \mathrm{N})$. 
\begin{table}[ht]
\caption{Error comparison between theoretical and equivalent models, in static and dynamic tests}
\centering
\setlength{\tabcolsep}{4pt}  % 缩小列间距
\begin{tabular}{lccc}
\toprule
\textbf{Metric} & \textbf{Static} & \textbf{Dynamic-Swing} & \textbf{Dynamic-Stance} \\
                & $\theta_1$ / $\theta_2$ & $\theta_1$ / $\theta_2$ & $\theta_1$ / $\theta_2$ \\
\midrule
RMSE (rad)   & 0.00096 / 0.00574 & 0.00094 / 0.00586 & 0.01693 / 0.03000 \\
MaxAE (rad)  & 0.00369 / 0.01452 & 0.00379 / 0.01559 & 0.04673 / 0.08805 \\
\bottomrule
\end{tabular}
\label{tab:error_summary}
\end{table}

\begin{figure*}[htbp]
\centering
\includegraphics[width=0.95\linewidth]{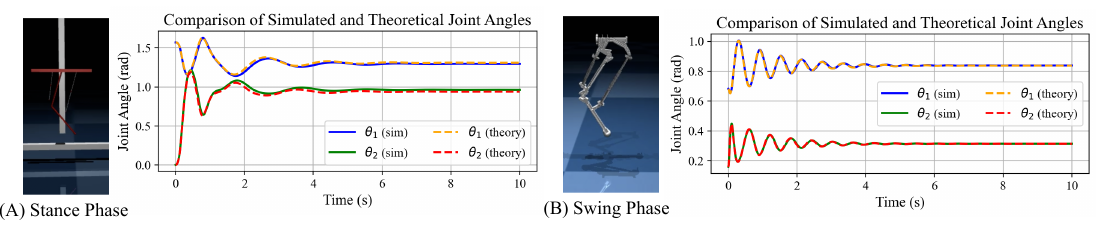}
\caption{Verification of dynamic equivalence in simulation. The stance phase (A) and swing phase (B) are tested. Joint trajectories from simulation and theoretical models show strong agreement, demonstrating the validity of the proposed formulation.}\label{fig-dynSS}
\end{figure*}
The error of both phases are summarized in Table \ref{tab:error_summary}, and trajectories of several experiments are visualized in Fig.~\ref{fig-dynSS}. The max absolute error (MaxAE) of all experiments of swing phase is less than $0.016~\mathrm{rad}\ (\sim 1~\mathrm{deg})$. The error of stance phase is about 5 times larger, which is still a small value compared to the workspace. The larger error may be caused by the relatively weak constraint between the foot and the ground. The results show that the equivalent model exhibits a small deviation from the theoretical model in step response, confirming the effectiveness and accuracy of the proposed equivalent modeling methods.

\subsubsection{Morphology Generalization}
\begin{figure}[htbp]
    \centering
    \includegraphics[width=0.95\linewidth]{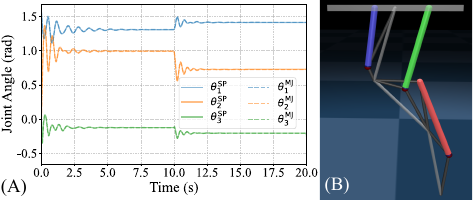}
    \caption{Morphology generalization on a 3-DOF musculoskeletal robot. (A) Joint-angle trajectories under a multi-pulse actuation sequence; solid: analytical (SymPy), dashed: EquiMus (MuJoCo). (B) Randomized 3-DOF morphology with three muscles highlighted.}
    \label{fig-3dof}
\end{figure}

We evaluated EquiMus on a randomized 3-DOF musculoskeletal robotic leg, varying muscle routing and joint-connection topology. Under identical inputs, time step, and solver settings, we applied a multi-pulse actuation sequence and compared joint trajectories from the analytical model (SymPy) against those from EquiMus, obtaining per-joint RMSEs of 0.0035, 0.0066, and 0.0041 rad, as illustrated in Fig.\ref{fig-3dof}. Those small errors validate that the energy-equivalent unit generalizes well to novel topologies, supporting the structural generalization of the framework.
\subsection{Physical Validation}
To make comparison of the simulation with real systems, parameter identification is conducted at both the actuator level and the system level, aiming to capture the key characteristics of individual actuators as well as the global static and dynamic behavior of the robotic leg.

\subsubsection{Actuator Calibration}
Actuator-level calibration was performed to determine parameters such as $m$ and $l_0$.

\subsubsection{Static Parameter Identification and Verification}
We tested the steady-state response with specific pressure commands. The data was recorded in the form $(\theta_1, \theta_2, P_1, P_2)$, representing the joint angles and corresponding actuator pressures at equilibrium. We used theoretical static model to calibrate static parameters, then calibration on $(k_1, k_2, s_1, s_2)$ became a linear regression problem.

In the experiment, the pressure was selected from 0--50~$\mathrm{kPa}$ at intervals of 10~$\mathrm{kPa}$, totaling 36 experiments (Set B). After excluding the experiments with antagonistic situations, 13 configurations $[P_1, P_2, \theta_1, \theta_2]$ are remained, forming Set A, which is used for system-level parameter identification. The results are listed in Tab.~\ref{tab:param_comparison}.    
The RMSE of pressure is 0.887~$\mathrm{kPa}$ for $P_1$ and 0.861~$\mathrm{kPa}$ for $P_2$, both below 2\% of the maximum applied pressure (50~$\mathrm{kPa}$), indicating a reliable regression accuracy.

\begin{table}[ht]
\centering
\begin{threeparttable}
\caption{Static validation results across two evaluation sets}\label{tab:static_error_rad}
\begin{tabular}{l@{\hspace{5pt}}c@{\hspace{5pt}}c@{\hspace{5pt}}c@{\hspace{5pt}}c}
\toprule
\textbf{Evaluation Set} & \textbf{RMSE $\theta_1$} & \textbf{RMSE $\theta_2$} & \textbf{MaxAE $\theta_1$} & \textbf{MaxAE $\theta_2$} \\
\midrule
Set A (13 poses) & 0.0170 & 0.0566 & 0.0336 & 0.1463 \\
Set B (36 poses) & 0.0203 & 0.0416 & 0.0418 & 0.1463 \\
\bottomrule
\end{tabular}
\begin{tablenotes}[flushleft]
\setlength{\labelsep}{0pt}
\setlength{\leftmargin}{0pt}
\footnotesize
\item \textbf{Note:} $E$ denotes error of pressure, expressed in kPa.
\item \textbf{Set A}: Used for parameter identification under moderate configurations.
\item \textbf{Set B}: Includes edge cases with joint limits to test robustness.
\end{tablenotes}
\end{threeparttable}
\label{tab:static_eval}
\end{table}

The parameters identified from Set A were then used to simulate the same pressure inputs for both configurations of Set A and Set B. The static joint angles were compared against experimental measurements. For Set A, compared to the actual positions, only one data point had an error exceeding 0.087 rad. This verifies the accuracy of the static model. As reported in Table \ref{tab:static_error_rad}, the model maintains low angular errors in both Set A and B. This confirms that our energy-equivalent modeling framework generalizes well, even in highly nonlinear and boundary-range conditions. Notably, larger errors in $\theta_2$ tend to occur when $\theta_1$/$\theta_2$ is near its joint limit. For example, MaxAE of $\theta_2$ in Set A and Set B occur simultaneously at $\theta_{1, 2}=2.068,0.18~\mathrm{rad}$, where $\theta_1$ is very close to its upper bound 2/3 $\pi$. In such poses, nonlinear effects caused by static friction and boundary constraints become more significant, making accurate simulation more challenging. While this introduces slight discrepancies, the overall predictive performance remains robust and suitable for downstream applications such as control and planning.
\begin{table}[htbp]
\centering
\caption{Comparison of statically and dynamically identified parameters}
\label{tab:param_comparison}
\begin{tabular}{lcc}
\toprule
\textbf{Parameter} & \textbf{Static Value} & \textbf{Dynamic Value} \\
\midrule
$k_1$ (N/m)        & 203.95                & 265.82                 \\
$k_2$ (N/m)        & 105.07                & 110.36                 \\
$s_1$ (m$^2$)      & 0.000411              & 0.000403               \\
$s_2$ (m$^2$)      & 0.000324              & 0.000436               \\
$l_{10}$ (m)       & 0.1642                & 0.1803                 \\
$l_{20}$ (m)       & 0.2579                & 0.2565                 \\
$c_1$ (Ns/m)       & –                     & 12.48                  \\
$c_2$ (Ns/m)       & –                     & 24.79                  \\
$c_{\text{hip}}$ (Ns/m) & –             & 0.524                  \\
$c_{\text{knee}}$ (Ns/m)  & –             & 0.062                  \\
\bottomrule
\end{tabular}
\end{table}

\subsubsection{Dynamic Calibration and Verification}
We tested step response in real world experiments and simulations to calibrate the main dynamic parameters $\mathbf{c} = \left[k_1, k_2, s_1, s_2, l_{10}, l_{20}, c_1, c_2, c_{\mathrm{hip}}, c_{\mathrm{knee}}\right]$. We tracked the actual joint trajectory $\tau_i(t)$ and the simulated trajectory $\hat{\tau}_i(t, \mathbf{c})$, respectively. We continued to use 13 groups got in last experiments. Finally we wanted to minimize the position error with the interpolation form of trajectory:
\begin{align}
    \mathbf{c} &= \arg\min\sum_{i=1}^{13} E(\mathbf{\tau}_i, \hat{\mathbf{\tau}}_i(\mathbf{c})).
\end{align}
Define the error $E(\theta_1, \theta_2)= \mathrm{MSE}(\tau(t)) + \lambda \mathrm{MSE}(\mathrm{sgn}(\dot{\tau}(t)))$ Here, $\lambda$ (set as 100 in experiments) is the weighting parameter that penalizes angular velocity signal error. The error represents the deviation in the system’s motion state $\left[\theta_1, \theta_2, \dot{\theta}_1, \dot{\theta}_2\right]$.

Our work employed differential evolution algorithm (DE) to do optimizations, with up to 10,000 iterations, and a population size of 25. Final parameters are shown in Table~\ref{tab:param_comparison}. Most of the parameters calibrated from the dynamic identification is closer to that calibrated from the static identification. The RMSE of the angle is under 0.03253~rad for all experiments. It indicates the validness of the equivalent method. The error may because of the nonlinearity of the actuator, which carry in some bias on the actuator dynamics.

\begin{figure}[htbp]
    \centering
    \includegraphics[width=0.99\linewidth]{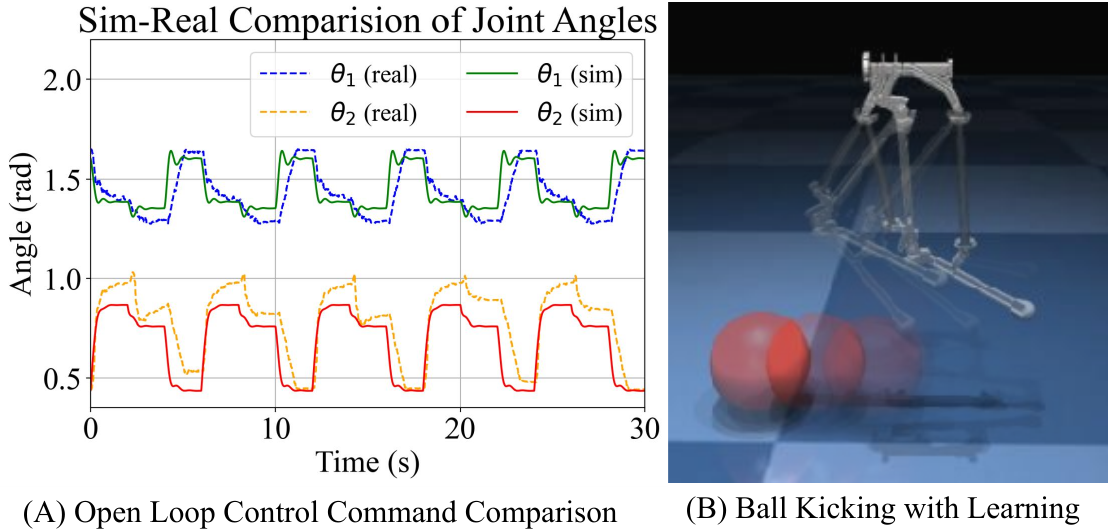}
    \caption{(A) Tracking performance of the triangular trajectory in both simulation and hardware experiments. Each vertex corresponds to a pressure command $P_{1,2}$: 
    (1) $6.15, 2.73$~kPa; 
    (2) $3.76,7.26$~kPa; 
    (3) $19.80,8.52$~kPa. (B) The robotic leg learns to kick a ball based on reinforcement learning.}
    \label{fig-triangle}
\end{figure}

In both MuJoCo simulation and real-world experiments, we commanded the robot to follow a triangular trajectory using a sequence of pressure steps. The resulting joint trajectories are shown in Fig.~\ref{fig-triangle}. They exhibit similar dynamic characteristics. The RMSE of $\theta_1$ and $\theta_2$ are 0.092 and 0.174~rad respectively. Some discrepancies between simulated and actual joint trajectories, especially in $\theta_2$, arise due to the joint's higher sensitivity to system disturbances. The presence of static friction may cause sticking or abrupt transitions and the effect of control latency in pneumatic valves may introduces slight time mismatches.

\subsubsection{Baseline Comparison with Native MuJoCo Implementation}
To benchmark the effectiveness of our energy-equivalent formulation, we constructed a baseline using MuJoCo’s native motor actuators configured as single spring-damper elements between bodies, without internal mass or constraint enforcement. The simulation was run under the same pressure inputs and mechanical boundary conditions as in our full model.
As shown in Table \ref{tab:baseline_comparison}, the baseline’s accuracy deteriorates significantly. These results underscore the necessity of incorporating intermediate mass elements and equality constraints to more accurately capture the actuator’s physical behavior.
This comparison highlights the advantages of our approach in preserving both simulation fidelity and real-time performance, making it well-suited for reinforcement learning and control applications.
\begin{table}[htbp]
    \centering
    \caption{Baseline comparison between
 EquiMus and native MuJoCo.}
    \label{tab:baseline_comparison}
    \begin{tabular}{lcc}
        \toprule
        Method & RMSE (rad) & Notes \\
        \midrule
        EquiMus     & 0.032522   & Energy-equivalent formulation \\
        PureMuJoCo  & 0.432144  & Native actuator, lumped model \\
        \bottomrule
    \end{tabular}
\end{table}
\subsection{Sensitivity Analysis of Model Parameters}
\begin{figure}[!tbp]
\centering
\includegraphics[width=0.98\linewidth]{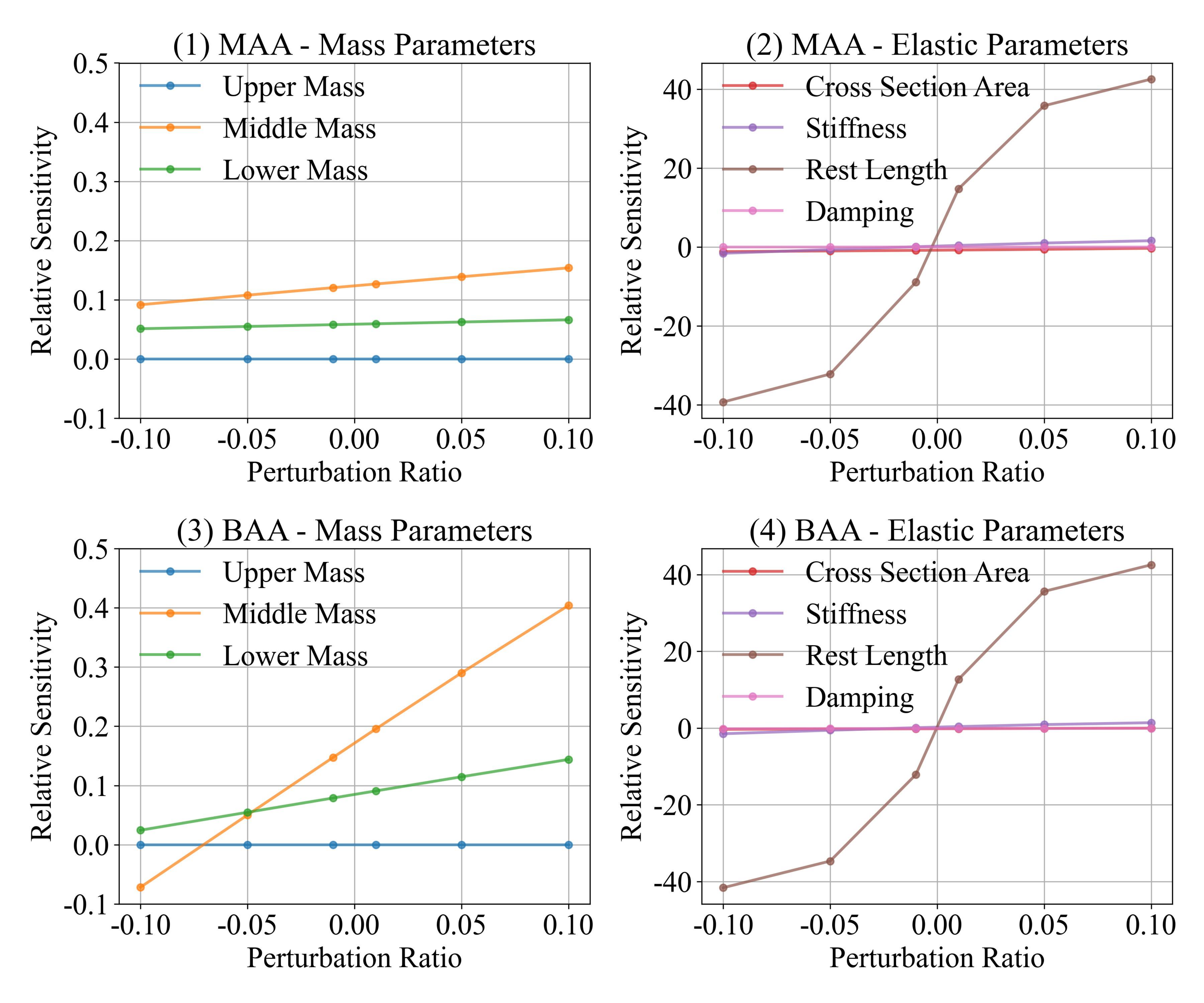}
\caption{Sensitivity analysis of the model parameters}\label{fig-sensitivity}
\end{figure}

To evaluate robustness under parametric uncertainty, we perturbed key actuator parameters ($m$, $k$, $l$, $c$, $s$) by $\pm1\%$, $\pm5\%$, $\pm10\%$ and measured the change in joint-angle RMSE with respect to real-world data. As shown in Fig.~\ref{fig-sensitivity}, the model remains robust under small perturbations, though rest lengths ($l_{10}$, $l_{20}$) and lower-link masses exhibit higher sensitivity. These critical parameters were precisely calibrated to ensure model fidelity during implementation.
\subsection{Application-Driven Demonstrations}
Built on MuJoCo, EquiMus leverages that ecosystem’s XML tooling, Gym-compatible interfaces, and mainstream RL/control libraries, enabling direct integration into standard pipelines with minimal additional code. We illustrated three representative usages. 
(i) \textbf{Reinforcement learning}: A ball-kicking task was achieved with PPO (Stable-Baselines3), with results shown in Fig.~\ref{fig-triangle} and the video. 
(ii) \textbf{Parameter exploration/identification}: Fast batched simulation supports sweeps of parameter like stiffness for system design and identification. 
(iii) \textbf{Model-based control}: Identified dynamics are deployable within MPC frameworks \cite{howell2022}. We also provide results of an auto PID tuning experiment at \href{https://github.com/fly-pigTH/EquiMus/tree/main/src/application/PID_AutoTuning/ReadMe.md}{\texttt{PID\_AutoTuning}}. These demonstrations indicate that EquiMus is not only physically consistent with hardware but also operationally practical for learning and control.

\section{Conclusion}

This letter proposed EquiMus, an energy-equivalent dynamics and simulation for the rigid--soft musculoskeletal robots with linear elastic actuators. By lumping actuator inertia with a 3–2–1 discretization, the method captures dynamic mass redistribution, supports loop-closure constraints in MuJoCo, and remains real-time capable. Experiments on a pneumatic leg show close sim-to-real agreement and enable downstream usage in PID auto-tuning, model-based control, and reinforcement learning. A current limitation is the omission of internal actuator nonlinearities; the modular energy-based structure readily admits richer constitutive models as future work. We hope EquiMus provides a novel perspective and serves as a small step toward bridging the gap between soft robotics and embodied intelligence.

\bibliographystyle{IEEEtran}
\bibliography{ref}

\end{document}